# Optimizing Violence Detection in Video Classification Accuracy through 3D Convolutional Neural Networks

Aarjav Kavathia*, Simeon Sayer
aarjavdkavathia@gmail.com, ssayer@fas.harvard.edu

*Primary Author

**Abstract**

As violent crimes continue to happen, it becomes necessary to have security cameras that can rapidly identify moments of violence with excellent accuracy. The purpose of this study is to identify how many frames should be analyzed at a time in order to optimize a violence detection model's accuracy as a parameter of the depth of a 3D convolutional network. Previous violence classification models have been created, but their application to live footage may be flawed. In this project, a convolutional neural network was created to analyze optical flow frames of each video. The number of frames analyzed at a time would vary with one, two, three, ten, and twenty frames, and each model would be trained for 20 epochs. The greatest validation accuracy was 94.87% and occurred with the model that analyzed three frames at a time. This means that machine learning models to detect violence may function better when analyzing three frames at a time for this dataset. The methodology used to identify the optimal number of frames to analyze at a time could be used in other applications of video classification, especially those of complex or abstract actions, such as violence.

## II. Introduction

There are roughly one security camera for every eight people, or a billion total [1]. Similarly large amounts of video content is uploaded online every day, with platforms such as YouTube handling 500 hours of videos uploaded every minute [2]. Therefore, there is an urgent need to scan, detect and alert this video content in order to keep both in person and online communities safe.

It is also therefore important that models are adapted to real world use cases, and can understand fundamental patterns of violence across datasets. Previous video classification models to detect violence, while impressive for what they were trying to achieve, have typically not been as applicable for real-world usage as the models have either relied on flawed data sets or outdated technology that means real time processing is slow. This has significantly inhibited their accuracy in real-world scenarios. Additionally, these previous techniques, such as acceleration differences between frames [3], may rely on correlation of violence within the dataset, rather than causation of violence. This paper argues that such methods are prone to overfitting, and do not allow the model to learn for itself violence specific macro structures.

To do this, this research explores the effect of the number of frames a convolutional network analyzes at a time and its effect on the validation accuracy of video classification. A 3D CNN was adjusted and optimized over the course of the experiment to 'look' at the right number of frames for context. Typically, prior video classification models examine two consecutive frames of a video in order to analyze the difference between those frames, such as changes in direction and speed of objects. This project analyzes the results of the multi-framed model as trained on a spectrum of 1, 2, 3, 10, and 20 frames at a time. The training data feature a widely used dataset consisting of hockey videos, which may or may not include fighting. The input data is a short hockey clip of about 40 frames with a width of 720 pixels and a height of 576 pixels (see Dataset). The output of the model results in a label for the video as either fighting or non-fighting. The accuracy of the model was determined using a test set, and the results analyzed with a confusion matrix to determine which type of inaccuracies were more prevalent. Additionally, an inspection of specific kernel values offered some insight into what the model is looking at when giving a violence determination. Although the model used here relies on recorded video rather than live footage, the results of this paper provides for better insights and understanding of how time, measured by number of frames, plays a role in violence detection models, with takeaways to improve real-time surveillance camera footage violence recognition.

The broader results of exploring the role of time are applicable across machine learning. As computer vision becomes more integrated into daily life, especially through the emerging capabilities of augmented reality, finding efficient and accurate ways to build models to identify complex human behaviors becomes increasingly important. Previous approaches have been to simply increase the size of kernels in traditional 2D models. However, complex actions often have a specific temporal fingerprint - that is, an action can not always be identified from a single frame. The findings of this research show a systematic approach to finding temporal analysis windows that allow for optimal multidimensional kernel training, and sheds insight into how future models could be built to facilitate detection of other actions.

**III. Background**

Since early computer vision research began, experiments have attempted to answer the question of what violence looks like. A 2011 Bermejo *et al* paper [4] titled "Violence Detection in Video Using Computer Vision Techniques" contains one of the earliest artificial intelligence models that focuses on classifying aggressive actions specifically, using a specially created dataset of hockey players either fighting, or playing normally. Other action recognition models at the time typically focused on basic actions such as clapping or jogging. Bermejo's model was able to distinguish between fighting and non-fighting videos with an accuracy of almost 90%. While this accuracy is impressive for the time, the model in the paper relied on some outdated technology such as the STIP and MoSIFT descriptors, as well as older and slower models, which this paper aims to improve upon. Deniz *et al's* paper "Fast violence detector" [3] attempts to approve upon the MoSIFT descriptor using a mathematical method of finding the acceleration exhibited between frames, achieving a 91.2% accuracy on the hockey violence dataset. While this model showed a significant accuracy improvement over the MoSIFT descriptor, it likely relies on specific quirks of the dataset. While change in acceleration clearly serves a useful proxy for violent action for their data, it does not necessarily find fundamental patterns or provide insight as to how violence manifests itself within data, which this paper hopes to explore. This also applies to key frame extraction techniques that apply 2D CNN's to the most important frames of the video, such as the 2023ConvLSTM paper [5]. While these models show high accuracy, there is no exploration of temporal significance which is important for continuous data streams, meaning these papers are arguably less applicable to live video feeds.

Working with different datasets, a 2018 paper [6] titled "Eye in the Sky" focused on using drones for violence detection in crowds and used deep learning to classify poses identified from images taken from the drone as either violent or nonviolent. The five violent poses were strangling, punching, kicking, stabbing, and shooting. The model reached a notable 94% accuracy, although it did drop to a still impressive 79% as more people entered the frame. This is a significant step in the abilities of AI and action recognition, but as the dataset was based on people pretending to attack each other, rather than actually doing so, the accuracy is questionable and the model may not be as applicable in real-life situations. Additionally, this relies on top down, well lit drone coverage, which is not always available. The posing and the lighting were especially important as the model relied upon the generation of skeletons for the humans in the image, from which the action was inferred. This introduces potential inefficiencies, as ideally the model itself would determine the important structures and features of the raw video in its classification.

There have also been significant commercial developments that demonstrate the validity of violence detection models, even if their methodology is not public. For instance, a company named ZeroEyes in Philadelphia [7] has developed and marketed technology to police that can detect concealed guns on a person. This system is useful as it could be a predictive indicator of gun violence. However, the accuracy may be questionable because, like the drones, the dataset comes from staged mock attacks. Additionally, the model appears to be focusing on an image of someone who possesses the weapon, rather than the actual act of violence itself, which means no analysis of temporal features are necessary.

Lastly, the company Athena, backed by famous investor Peter Thiel, has developed software to recognize aggressive actions such as fighting or walking fast based on a combination of the video and a separate model that analyzes a person's emotions and facial expressions [7]. This innovative model claims an accuracy of 99%. However, the details of the methodology are not public, and therefore the broader lessons of how models process complex actions over time are not yet fully explored. This paper intends to build out detection algorithms using publically available data, to which it will apply custom CNN architectures in order to both create an open-source violence detection system, and as an experiment that explores complex action detection in video data through the example of violence detection.

## IV. Dataset and preprocessing

The dataset used for this model was created by the aforementioned paper, "Violence Detection in Video Using Computer Vision Techniques" [4] and was made public by the original authors. It consists of 1000 video clips taken from professional ice hockey games, with a length of approximately 2 seconds, in which either a normal game sequence was shown, or a fight between two players. The video clips were compiled from the National Hockey League. They were originally of resolution 720 X 576 pixels, but due to data restrictions, the length and width of the video clips were each shrunk to quarter size to yield video dimensions of 180 X 144 pixels.

The model represents this data therefore as a four dimensional array, with a width of 180, a height of 144 and an effective depth of 40, for the frames. This is all extended by an additional dimension to account for the RGB values of each of the pixels.

The final dataset consisted of a randomly selected 200 videos of the original 1000 videos, to ensure efficient run times. Of the 200 total videos, 100 were labeled as fighting and 100 were labeled as non-fighting.

A still frame of a fighting video is shown below in figure 1:

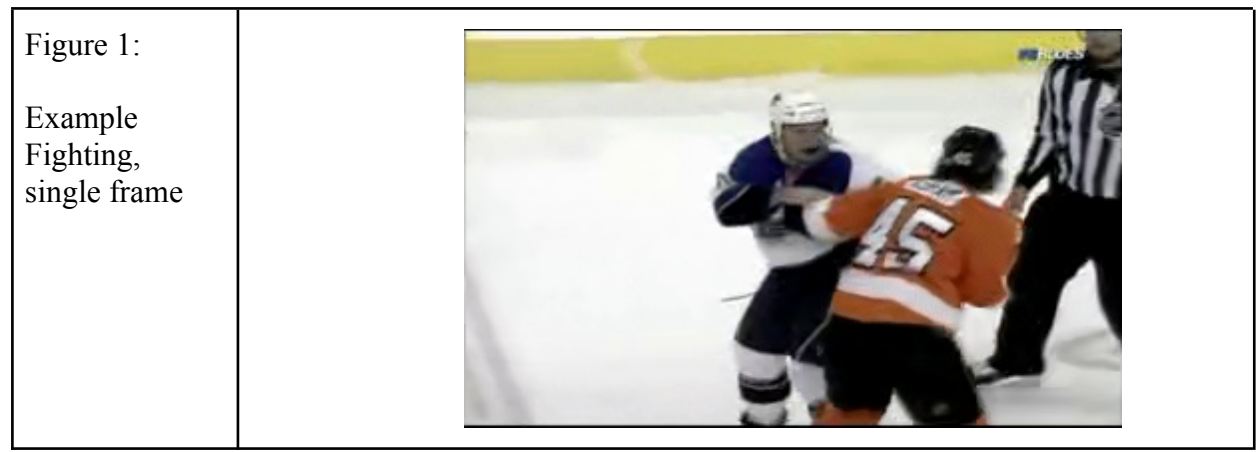

| Figure 1: Example Fighting, single frame | |

The video, as shown here, includes logos, numbers, and other graphics associated with the NHL broadcast of the game. It also may include other players, referees or equipment as part of the clip.

Each of the images that make up the 40 frames in a given video were processed and translated into an optical flow image, using a python implementation of the Lucas-Kanade method [8] to determine the overall motion of the video. The specific type of color represents the direction of the motion in the frame while the intensity of the color would represent the magnitude of the motion. An example is shown below in Figure 2:

| Figure 2: Side by Side comparison of Optical Flow representation and its original frame | 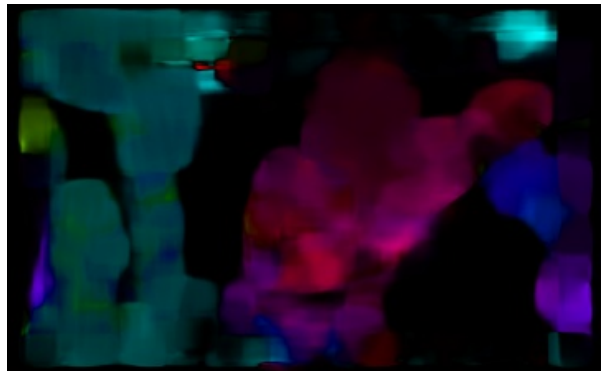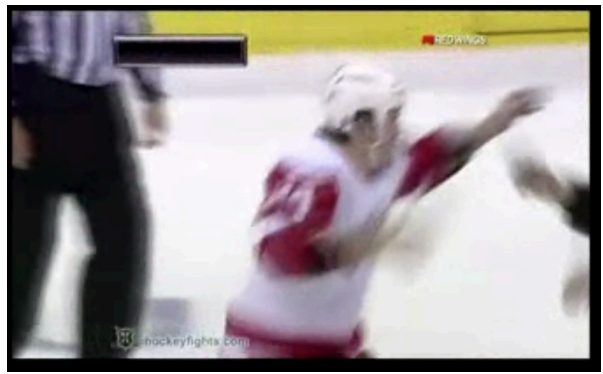 |
|---|---|

The advantages of running the model on the optical flow dataset is that certain non-useful elements of the image are removed. Because the logos are static, they do not appear in the optical flow data. It also allows a single frame to have some inherent data encoded within it associated with the frames before. For instance, as seen in Figure 2, even a still frame now shows the direction in which the player is coming from, categorized by the red color that shows they are approaching from the left, moving to the right. This provides more useful information about motion and flow to the model, which assists in classification of violence, and for clearer understanding of temporal significance in action detection.

**V. Methodology/Models**

The structure of the machine learning model relies on the preprocessing described in the dataset section - that is, the conversion of the hockey video frames to an optical flow representation. This is then fed into the model, at its original size and with the 3 RGB color channels. It is then shrunk using a max pool, followed by the application of 6 convolutional kernels. The size of these layers, and their corresponding results are the study of this paper. Finally, the results of those kernels are fed to a perceptron of size 16, before an output layer determines the final prediction. The following figure illustrates this structure for the case of a 3x3x3 convolutional size.

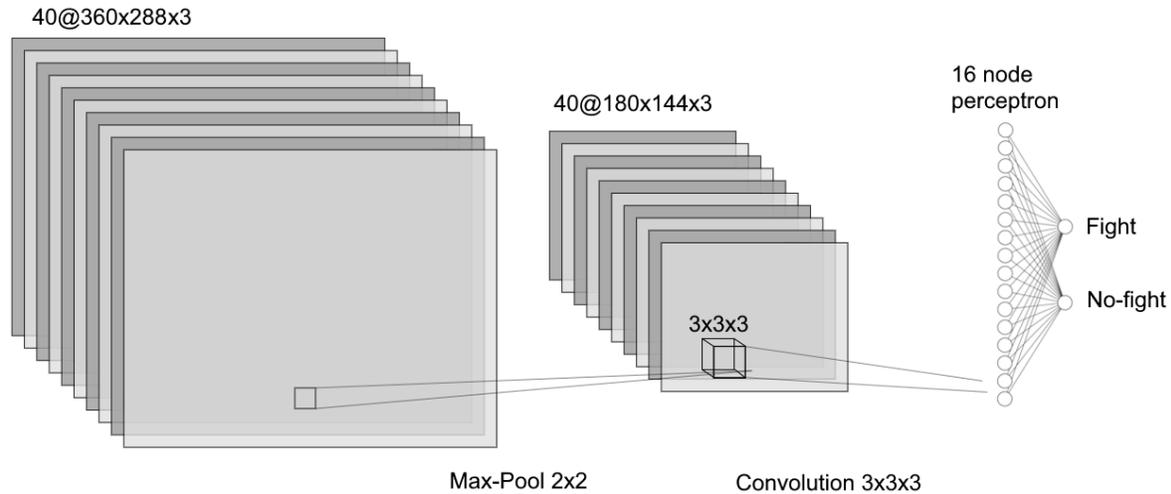

**Fig. 3: Example model structure for 3x3x3 CNN**

Various design considerations are embedded in the structure of the model. The primary feature of this model is the 3x3xN convolutional layer that takes a 3x3 reading of N frames of the data, allowing the model to train on how the image looks in a given frame, and changes over the N frames it has access to. This effectively allows us to control how long a window the model has to analyze the movement of hockey players and the corresponding optical flow. The original video was taken at 24 frames per second, meaning that each additional frame available to the convolutional, that is as N increases by 1, allows the model to train and find the relationship between a given color channel's data for an additional ~0.04 seconds of the footage in its determination of whether or not that video contains fighting. Having access to all color channels is important, as the optical flow data color is significant in describing the direction of movement within the image. Further research is needed to understand the relationship between kernel depth and size, and whether optimal configuration differs between actions.

The overall effect of this model structure is that the kernels are simultaneously able to capture the direction, speed, acceleration and interaction of the fighting, and how all of those properties change over time. This multidimensional approach is significantly more computationally expensive, but inherently provides more interesting findings. For instance, by adjusting the value of N and observing the changes within the model structure and the accuracy of the model, conclusions can be drawn about whether violence itself is more a function of space, or whether it is more linked to how that space changes over time.

The model was trained on a variety of structures using the prepared data. 20% of the data was kept for testing the model, to allow for effective testing against overfitting, which was especially significant. The dense amount of kernels and unique information able to be extracted from any given video meant that this model was especially prone to overfitting. Adding more data mitigates this somewhat, but comes at increased computational costs. The primary way to ensure that the model did not overfit was to keep the perceptron size small, to allow for general patterns to be conveyed, but to minimize the number of weights connected to the output layer which may have allowed for unique encodings of each video. A RELU activation function was used for the perceptron, and the optimizer deployed in the training of the model was Adam. This training occurred across all values of N.

## VI. Results and Discussion

The model was trained in a variety of network structures, as the value of N frames within the CNN changed. Figure 3 and Figure 4 (below) show graphs of a sample of the results.

| Figure 4: Epoch vs Validation Accuracy Graph | 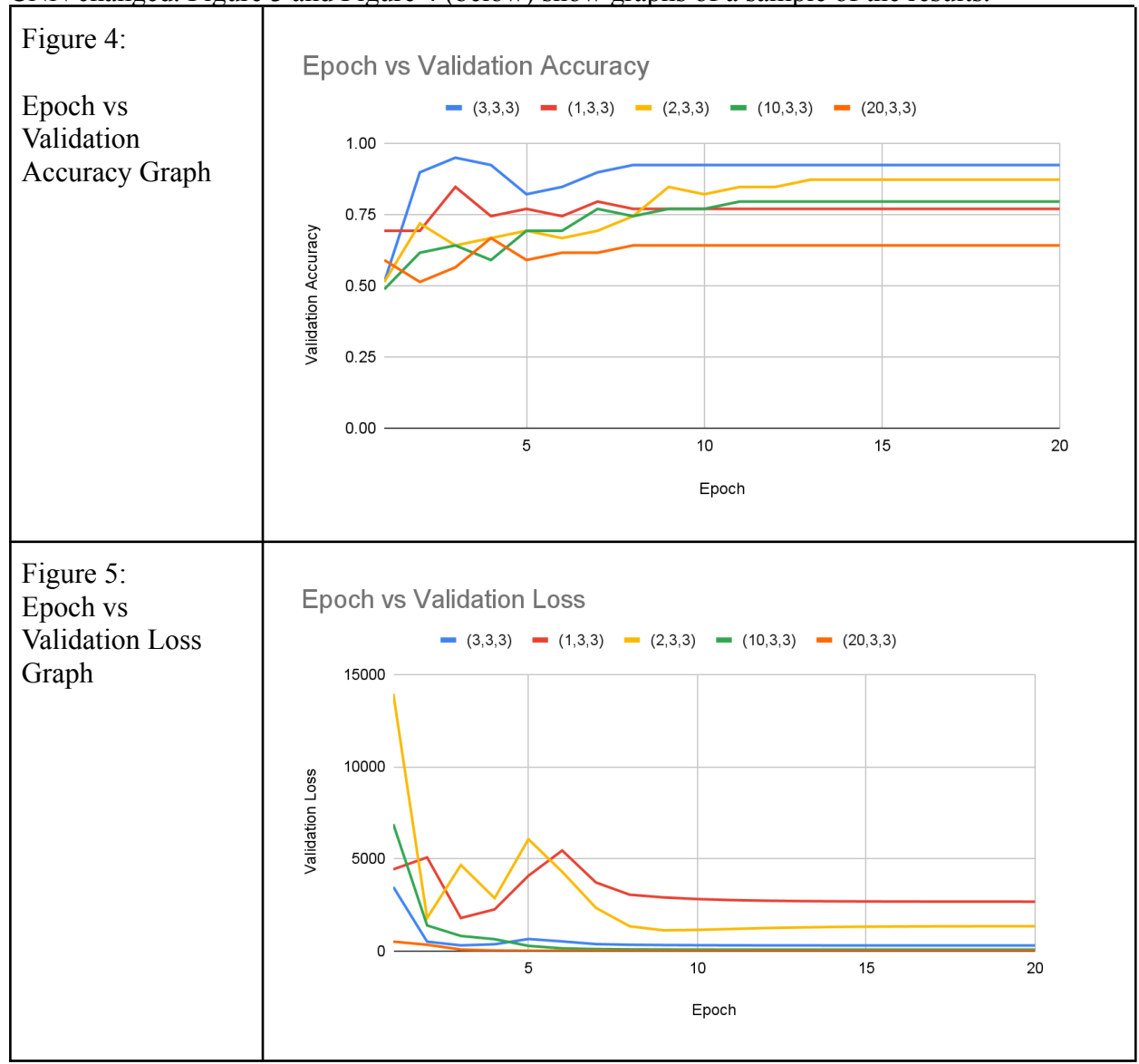 |
|---|---|
| Figure 5: Epoch vs Validation Loss Graph | |

After training and testing the model to analyze 1, 2, 3, 10, and 20 frames at a time, the model of kernel size 3x3x3, examining 3 frames at a time, yielded the greatest validation accuracy, as can be seen in the Epoch vs Accuracy graph in Figure 3, resulting in a peak validation accuracy of 94.87%, which eventually stabilized at 92.31% (Fig. 4), an improvement on many previous models. The confusion matrix for the 3x3x3 model (Fig 6.) shows that the errors made by the model were balanced, indicating the model's reasoning for its classification of violence is nuanced.

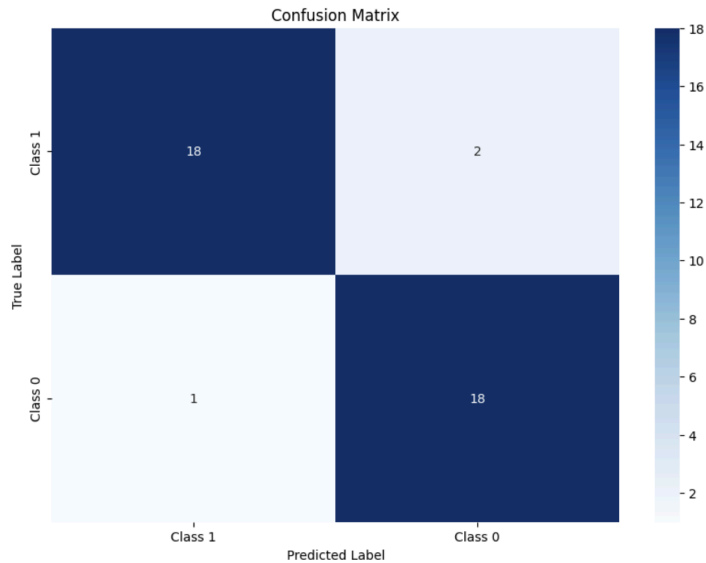

Figure 6. Confusion Matrix for the (3,3,3)

Interestingly, the (1,3,3) model which only analyzed one frame at a time reached a peak accuracy of 84.62% and plateaued at 76.92%. This demonstrates the importance of temporal features and multi dimensional kernels. One frame does not capture enough data for the model to learn the specific signs of violence in the video to classify violence correctly.

The paper by Deniz [3] utilized a 'FAST' model that analyzed the changes in acceleration between 2 frames, which is therefore comparable to a (2,3,3) model, which also analyzes two frames. It achieved an accuracy of 91.2% whereas the (2,3,3) model in this paper stabilized at a slightly reduced maximum validation accuracy of 87.2%. However, the (3,3,3) model which achieved a maximum validation accuracy of 94.9% did marginally better than the Deniz paper when analyzing the hockey violence dataset, showing the importance of spatial-temporal features that 3 frames offers. However, it seems that analyzing too many frames can hinder the accuracy of the video classification model. For example, the model that looked at every 10 frames stabilized at a maximum validation accuracy of 79.5% while the model that looked at every 20 frames achieved 66.7%. Having this larger window also seems to have contributed to overfitting by the model, which is logically more likely to occur as the model has access to more weights, despite those weights only being represented by 16 neurons. However, the fact that a 10 frame model was able to extract useful features to produce a 79.5% accuracy, nearly as accurate as a 1 frame model, indicates that longer temporal resolution can still produce useful results, which may be further refined with access to more training data.

This is confirmed by investigating the model weights. Some degree of interpretability is offered when analyzing the specific kernel values generated by the model, shown below in Figure 7. While it is generally understood that CNN's are black box, and therefore not optimal for inferring specific meaning, by understanding the optical flow data, it is known that red and pink colors represent objects moving from right to left, and blues objects moving from left to right. Therefore, this kernel seems to activate when it detects a 'clash' between two objects moving against each other, as would be expected in a fight. Further research could be taken to better understand how models think about shapes moving over time, and its ability to recognize actions within video data.

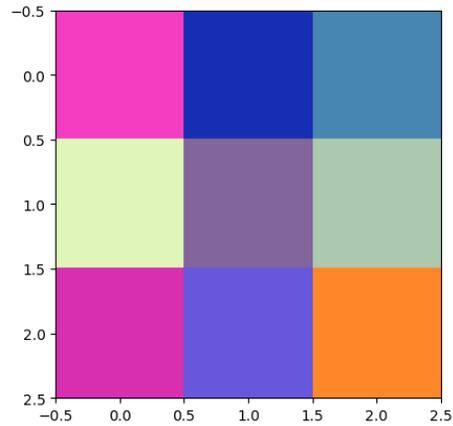

Figure 7. Model Weight Slice

The models therefore performed at or above existing models, while using comparatively more simple model structures. There are broader lessons that can be applied from this experimentation, and more research to be done on both this specific violence detection model, and on action recognition from video using multi dimensional CNN's as a whole.

**VII. Conclusion**

The primary goal of this paper was to explore a methodology to determine the optimal number of frames to be examined by a video classification model at a time in order to obtain the greatest accuracy. At first, each video in the dataset was split into frames which were then converted into optical flow images. Ultimately, the video classification model that analyzed three frames at a time yielded the best results with a peak testing accuracy of 94.87% when detecting violence in hockey videos. The results of this paper mean that video classification models that analyze three frames at a time appear to yield the best results for this dataset. The broader methodology used to determine this could therefore be utilized in commercial applications that involve video classification. Some next steps for optimizing violence detection models could be obtaining and training on more violence data from movies and real-life situations to make the model more accurate when used in a commercial security camera.

# VIII. Acknowledgements

The author would like to acknowledge the guidance of Simeon Sayer from Harvard University for his mentorship on the methodology and implementation of the video classification model throughout this project.